\documentclass[journal]{IEEEtran}

\usepackage[utf8]{inputenc}
\usepackage[T1]{fontenc}
\usepackage{amsmath,amsfonts,amssymb}
\usepackage{xcolor}
\usepackage{graphicx}

\usepackage{balance}
\usepackage{cite}
\DeclareMathOperator{\atantwo}{atan2}
\usepackage{subcaption}

\newcommand{\R}{\mathbb{R}}

\usepackage{tikz}
\usepackage{tikzscale}
\usetikzlibrary{arrows}
\usetikzlibrary{external}
\usepackage{pgfplots}
\usepgfplotslibrary{external}

\pgfplotsset{
    compat = newest,
    legend image with text/.style={
        legend image code/.code={%
            \node[anchor=center] at (0.3cm,0cm) {#1};
        }
    },
}
\tikzset{every picture/.append style={font=\normalsize}}

\DeclareMathOperator*{\argmin}{argmin}   

\begin{document}
\title{Recalibrating the KITTI Dataset Camera Setup for Improved Odometry Accuracy}
\author{Igor~Cvišić, Ivan~Marković and~Ivan~Petrović$^\ast$
%
\thanks{$^{\ast}$Authors are with the University of Zagreb
Faculty of Electrical Engineering and Computing, Laboratory for Autonomous Systems and Mobile Robotics, Croatia. {\{igor.cvisic, ivan.markovic, ivan.petrovic\}@fer.hr.}\\
\indent This research has been supported by the European Regional Development Fund under the grant KK.01.1.1.01.0009 (DATACROSS).

\vspace{2mm}
\noindent 978-1-6654-1213-1/21/\$31.00~\copyright2021 IEEE
}}


\maketitle

\begin{abstract}
Over the last decade, one of the most relevant public datasets for evaluating odometry accuracy is the KITTI dataset. Beside the quality and rich sensor setup, its success is also due to the online evaluation tool, which enables researchers to benchmark and compare algorithms.
The results are evaluated on the test subset solely, without any knowledge about the ground truth, yielding unbiased, overfit free and therefore relevant validation for robot localization based on cameras, 3D laser or combination of both.
However, as any sensor setup, it requires prior calibration and rectified stereo images are provided, introducing dependence on the default calibration parameters.
Given that, a natural question arises if a better set of calibration parameters can be found that would yield higher  odometry accuracy.
In this paper, we propose a new approach for one shot calibration of the KITTI dataset multiple camera setup.
The approach yields better calibration parameters, both in the sense of lower calibration reprojection errors and lower visual odometry error.
We conducted experiments where we show for three different odometry algorithms, namely SOFT2, ORB-SLAM2 and VISO2, that odometry accuracy is significantly improved with the proposed calibration parameters.
Moreover, our odometry, SOFT2, in conjunction with the proposed calibration method achieved the highest accuracy on the official KITTI scoreboard with 0.53\% translational and 0.0009 deg/m rotational error, outperforming even 3D laser-based methods.
\end{abstract}

\begin{IEEEkeywords}
KITTI dataset, camera calibration, stereo camera, visual odometry.
\end{IEEEkeywords}

%

\section{Introduction}
\label{sec:intro}

Autonomous operation of mobile robots and vehicles is founded on processing signals from a suite of different sensors.
Such a suite can contain imaging sensors, such as color, grayscale, omnidirectional or thermal cameras, ranging sensors, such as sonars, radars and lasers, and proprioceptive sensors, such as inertial navigation systems coupled with GNSS.
The sensor suite needs to be calibrated extrinsically, i.e., we need to be able to determine relative poses between all the sensors and the robot or vehicle frame.
Unlike ranging sensors, which report depth by default, standard cameras loose this information in the imaging process.
Therefore, in order to be able to provide geometrical inference with respect to the environment and correct for non-linearities in the optical system, they need to be calibrated both intrinsically and extrinsically.
The missing depth can be alleviated by adding a second, or multiple cameras, in order to conduct stereoscopic reconstruction.
In this case, the prerequisite of accurate calibration also applies, as it will directly affect the accuracy of estimated depth which can later by used by various algorithms, such as visual odometry and simultaneous localization and mapping (SLAM).

For calibrating a single camera, probably the most popular approach is the Matlab toolbox \cite{Bouguet}, also being available under OpenCV \cite{Bradski2000}.
This approach is based on the pinhole camera model, enables intrinsic camera calibration, includes different lens distortion models, and also extrinsic calibration of a stereo camera.
The calibration is carried out using a known calibration pattern, most often the checkerboard pattern, since it facilitates detection of salient features like corners, and resolves the scale ambiguity by knowing the inner corner distances.
Reprojection error, i.e., the measure of distances between the detected and model reprojected corners in the image, is used as the optimization criterion.
To achieve best results, corners need to be localized with high sub-pixel accuracy and corner detection plays an important role in calibration algorithms.
In \cite{Wang2010} for automatic calibration from multiple images, authors propose to first detect image corners and then perform recognition of corners at the intersections of black and white squares and at the intersections of two groups of grid lines.
A related problem is also that of detecting and decoding visual fiducial tags, e.g., the AprilTag 2 algorithm \cite{Wang2016a}.
A popular framework is the Kalibr toolbox \cite{Furgale2013a, Maye2013, Furgale2015} offering calibration of multiple cameras, visual-inertial units, and rolling shutter cameras.
Its pattern is based on AprilTags, while circle and ring planar calibration patterns can also be used \cite{Datta2009}.

An important instrument in advancing fundamental methods for robot and vehicle autonomy are public datasets as they enable evaluation and comparison of different approaches.
Regarding visual odometry and SLAM, several datasets have been published over the years in the robotics and vehicle domain: the KITTI dataset \cite{Geiger2012CVPR,Geiger2013}, Málaga Urban dataset \cite{Blanco-Claraco2013}, KITTI-360 dataset \cite{Xie2016a}, The EuRoc micro aerial vehicle dataset \cite{Burri2016}, Oxford Robotics Car dataset \cite{Maddern2017}, Multivehicle Stereo Event Camera Dataset (MVSEC) \cite{Zhu2018a}, and a Stereo Event Camera Dataset (DSEC) \cite{Gehrig2021a}.
Regardless of the dataset, a visual camera setup requires prior calibration -- ideally for each sequence when outdoor recording is performed in diverse weather conditions.
All datasets usually provide camera calibration parameters, but sometimes the dataset also includes the calibration images enabling researchers to conduct calibration by themselves.
Such is the example of the KITTI dataset, which has been acting as public odometry and SLAM benchmark for road vehicles since 2012.\\

Contrary to the classical approach where mutiple images of the calibration pattern, i.e. boards, are taken, the KITTI was calibrated with a single shot of many boards, which is very practical when calibration needs to be performed for each recorded sequence.
However, this also presents a disadvantage of having smaller variability in the distance and orientation of the calibrations boards.
Since visual odometry accuracy is directly dependent on the calibration, having more accurate parameters can improve the overall performance.
In \cite{Kreso2015} authors proposed to calibrate a discrete stereo deformation field above the two rectified image planes, which would be able to correct deviations of the real camera system from the default calibration model.
The authors showed that the calibrated deformation field improves the accuracy of the recovered camera motion.

In this paper we propose a novel approach for one shot calibration of KITTI cameras.
Instead of directly detecting corners, we detect line segments within a RANSAC procedure and once the board geometry is recovered and neighboring edges are found, precise corner position is computed with sub-pixel accuracy using line intersections.
Thereafter, board matching is performed and the proposed method shows smaller reprojection error than the method in \cite{Geiger2012} and OpenCV.
To further enhance the odometry accuracy, we also perform grid search optimization for 4 out of 24 parameters for which the configuration of the boards did not strongly constrain the optimization problem.
We tested three methodologically different odometries, namely our SOFT2 \cite{Cvisic2021a}, ORB-SLAM2 \cite{Mur-Artal2016}, and VISO2 \cite{Geiger2011} -- all three showed higher accuracy with the proposed calibration parameters.
Moreover, our SOFT2 with the proposed calibration method scored 0.53\% in translation error and 0.0009 deg/m in rotation error rendering it currently the highest ranking algorithm on the KITTI scoreboard\footnote{http://www.cvlibs.net/datasets/kitti/eval\_odometry.php}.
Although the method is focused on a single dataset, we believe it to be relevant to the community due to its general insights, especially when benchmarking over odometry and SLAM with other sensors such as 3D laser, and relevance that KITTI had and probably will in the forthcoming years.

\section{Default KITTI Calibration Procedure}\label{sec:kitti_origigi}

As stated earlier, camera calibration is usually computed from images of a calibration board that has easily detectable patterns of known size, e.g., the checkerboard pattern.
The images should capture the board at different orientations and distances, and larger the diversity of board poses, the more constrained optimization with the camera-lens model is.
Otherwise, ambiguity emerges, since the system can be equally well represented with a different set of parameters.
Contrary to this one-board-many-images approach, KITTI was calibrated with a many-boards-one-image approach \cite{Geiger2012} using \emph{libcbdetect}.
This method is very practical, especially if one needs to repeat the calibration process every day, as was the case during KITTI dataset recording, but it also has several disadvantages.
First, the number of boards that can fit into a single image is limited.
The odometry dataset was calibrated with $12$ boards.
Then, since all $12$ boards reside in the same image they all are moderately far away from the camera and there cannot be a single board in the close view that would impose several constraints with high signal-to-noise ratio, since it would occupy a large part of the image.
Second, all squares are relatively small, resulting in higher corner position uncertainty relative to the square size, and board orientations are biased, i.e., all boards are tilted towards the image center.
For example, switching the positions between left and right boards without changing their angles would impose some new constraints and the same applies to the top and bottom boards.
Third, larger part of the image is not covered with boards, which could easily be avoided with standard calibration procedure.

Once the board images are acquired, all board poses relative to the camera are found together with intrinsic and extrinsic camera parameters.
KITTI was calibrated with the widely used radial-tangential model for lens distortion.
Given that, the next step is to find 2D corner locations, and for the KITTI dataset in \cite{Geiger2012} authors used corner templates -- one for horizontal boards and the other for boards rotated at 45 degrees.
In essence, any method can be used as long as it detects all the corners, but special attention needs to be paid to the sub-pixel refinement of corner locations.
Calibration parameters are very sensitive to changes in corner positions and having accurate corner locations before optimization process is of the utmost importance.
KITTI corner refinement is based on the fact that at a corner location $\mathbf{c} \in \R^2$ the image gradient $\mathbf{g_p} \in \R^2$ at a neighboring pixel $\mathbf{p} \in \R^2$ should be approximately orthogonal to $\mathbf{p}-\mathbf{c}$, leading to the optimization problem
\begin{equation}
\mathbf{c}=\argmin_{\mathbf{c'}}{\sum_{\mathbf{p} \in \mathcal{N}_{\mathbf{I}}(\mathbf{c'})}\
	{(\mathbf{g}^T_\mathbf{p}(\mathbf{p-c'}))^2}},
\end{equation}
where $\mathcal{N}_{\mathbf{I}}$ is a local pixel neighborhood around the
corner candidate.
Authors solved this problem in closed form:
\begin{equation}
\mathbf{c}=(\sum_{\mathbf{p} \in \mathcal{N}_{\mathbf{I}}}{\mathbf{g_p}\mathbf{g}^T_\mathbf{p}})^{-1}\
\sum_{\mathbf{p} \in \mathcal{N}_{\mathbf{I}}}{(\mathbf{g_p}\mathbf{g}^T_\mathbf{p})\mathbf{p}}.
\end{equation}
Corner refinement in the \emph{OpenCV} \cite{OpenCV} implementation uses function \emph{cornerSubPix()}, which is based on the same observation.
Indeed, in Table \ref{tbl:rep_err_compare} we can notice similar overall reprojection errors for \emph{libcbdetect} and \emph{OpenCV}.

However, refinement method used in \emph{libcbdetect} and \emph{OpenCV} does not perform optimally in the case of KITTI calibration images and there are several possible reasons behind that.
First, due to non-diffuse lightning, some boards are overexposed resulting in white squares being slightly bigger than the black ones.
Opposite edges of the corner are not exactly on the same line and the observation on which the refinement solution is based is violated.
Second, all boards are relatively distant from the camera and squares appear small having the side size of only 20 or even 6 pixels.
That explains maximum refinement window size of $9\times 9$ pixels we obtained in our experiments, which is close to the $11\times 11$ mentioned in \cite{Geiger2012}.
It would seem that signal-to-noise ratio for this neighborhood size is the limiting factor for more accurate corner localization.

\section{Proposed KITTI Stereo Camera Calibration}

In this section we present our novel calibration procedure for a stereo camera setup.
Compared to the default KITTI method and OpenCV's method it yields smaller reprojection errors.
We first describe how corners are detected and positions refined, followed by board matching and optimization.
Finally, a subset of parameters is refined using grid search to yield final values that increase the accuracy of the three tested visual odometry algorithms.

\subsection{Corner detection and refinement}

Our method is based on a different approach than the default KITTI calibration \cite{Geiger2012}.
Instead of detecting corners directly, we detect line segments targeting the sides of the board squares similarly to AprilTag 2 \cite{Wang2016a}.
Once the board geometry is recovered and neighboring edges are found, accurate corner location is computed as intersection of two lines.
The whole process starts with a robust Canny edge detector.
First, the image is converted to floating point and all subsequent operations are performed in floating point to minimize data loss.
Image is smoothed with a Gaussian filter of $5\times5$ pixels and the smoothed image is convoluted with a Sobel filter, producing image derivatives in horizontal $G_x$ and vertical $G_y$ direction.
From this, the edge gradient and direction can be determined
\begin{equation}
G=\sqrt{G_x^2+G_y^2},
\end{equation}
\begin{equation}
\Theta=\atantwo{(G_y,G_x)}.
\end{equation}

\begin{figure}[!t]
	\centering
	\includegraphics[width=0.45\textwidth]{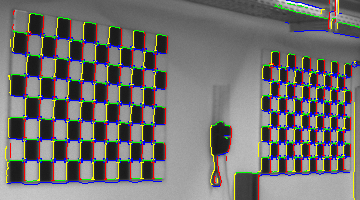}
	\caption{Detected edge pixels in the left image 0 from the KITTI calibration sequence 2011-10-03. Positive vertical edges are shown in red, negative vertical in yellow, positive horizontal in blue, and negative horizontal in green color.}
	\label{fig:lines_segment}
\end{figure}

The gradient image is filtered first with absolute suppression and then with non-maximum suppression.
All remaining edge pixels are localized with sub-pixel precision and correlated with corresponding direction.
According to the direction, edge pixel is assigned to one of the four possible classes: horizontal positive, horizontal negative, vertical positive, and vertical negative.
Positive and negative edges refer to the black-white order of squares in the corresponding direction.
Fig.~\ref{fig:lines_segment} shows detected edge pixels in the segment of the left image 0 from the KITTI calibration sequence 2011-10-03.
Thereafter, adjacent pixels belonging to same class are grouped into segments, each representing one exact side of the checkerboard square containing the subpixel location of each edge pixel belonging to that side.
Now, the task of finding a corner becomes a task of finding four different classes of segments with their ends very close to each other.
To reduce the number of outliers, only segments with similar lengths are connected and after creating a graph of connections to neighboring segments the board is reconstructed.
The first corner is always the one with exactly two connections: one to the right and one downwards.
First, we start moving to the right, adding new corners until we reach the corner without its right connection.
At this point, if a downward-connection exist, we move down and then to the left, following a snake-like pattern until we reach the ending corner without down-connection.
At the end of each row, we compare number of corners to the previous row and discard the current pattern if these numbers do not match.
This simple check filters out remaining outliers emerging from sources other than the board.

Finally, corner location is refined as follows.
Two equations of the line are found for each corner, one passing through two horizontal neighboring segments and the other one passing through two vertical neighboring segments.
For each line, a RANSAC procedure is used to suppress outliers, where an inlier is accepted if both the distance to the line is below the threshold and the point gradient direction is perpendicular to the line within some angle tolerance. After the line equations are computed, the precise corner location is found as the intersection of these two lines.

\begin{table}[!t]
	\caption{Reprojection error comparison between our method, \emph{libcbdetect}, and OpenCV's method. Last column shows result without board number 11, which is bent and therefore introduces larger error into optimization procedure}
	\label{tbl:rep_err_compare}
	\begin{center}
		\begin{tabular}{ |c|c|c|c|c|  }
			\hline
			& libcbdetect & OpenCV (9x9) & ours & ours wo b11\\
			\hline
			date & r. err. [pix] & r. err. [pix] & r. err. [pix] & r. err. [pix] \\
			\hline
			09-26 & 0.117022 & 0.116382 & \textbf{0.095050} & 0.066126 \\
			09-28 & 0.128945 & 0.122802 & \textbf{0.100227} & 0.072588 \\
			09-29 & 0.135855 & 0.128055 & \textbf{0.106766} & 0.074046 \\
			09-30 & 0.135186 & 0.127309 & \textbf{0.105401} & 0.070892 \\
			10-03 & 0.138393 & 0.129011 & \textbf{0.105318} & 0.073021 \\
			\hline
		\end{tabular}
	\end{center}
\end{table}

\begin{figure*}[!t]
	\centering
	\includegraphics[width=0.8\textwidth]{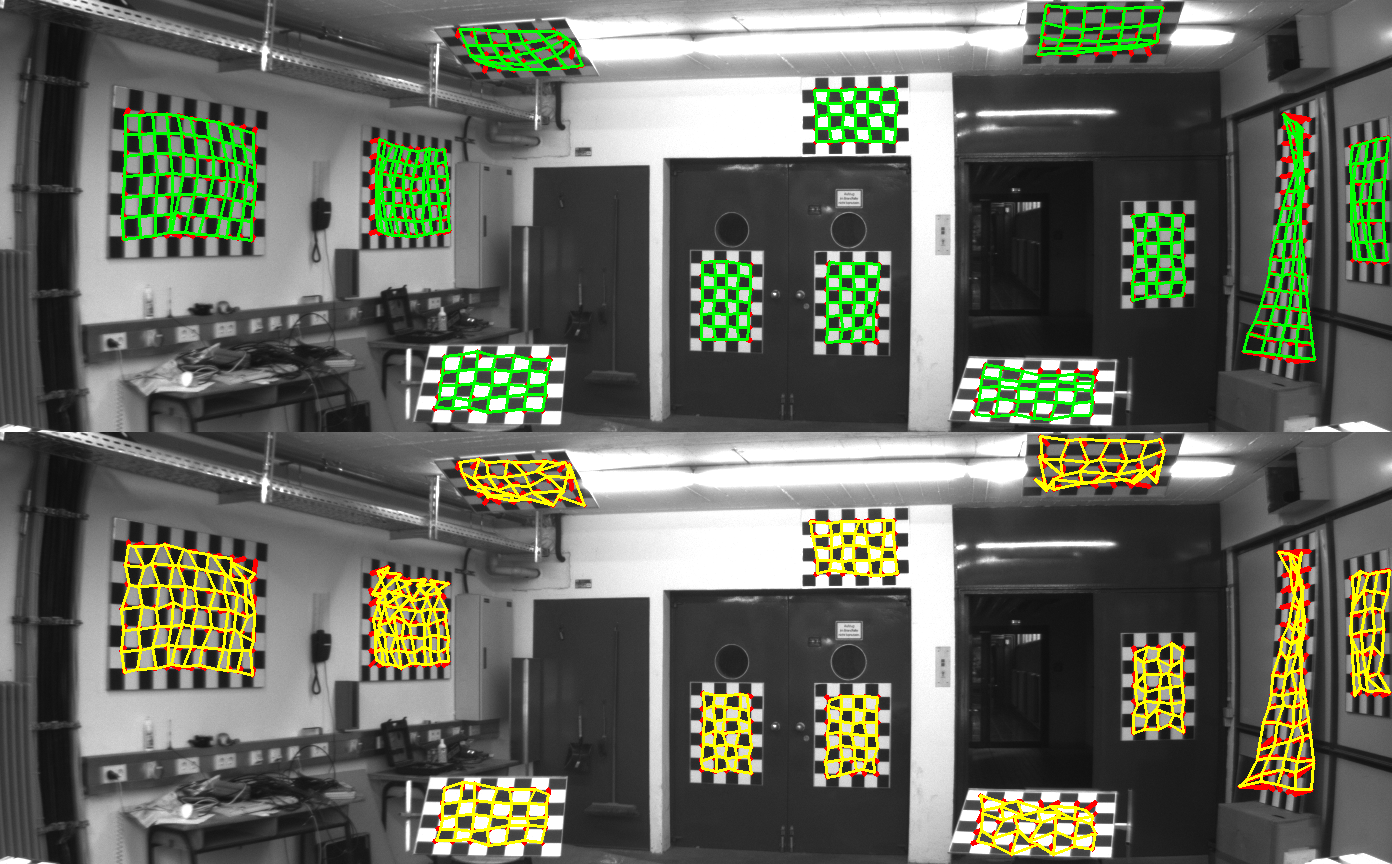}
	\caption{Reprojection error vectors for the proposed line intersection method (up, green) and \emph{libcbdetect} (down, yellow) magnified $50\times$ for the left frame 0 of KITTI calibration sequence 2011-10-03}
	\label{fig:reprojection_error_comparison}
\end{figure*}

Fig.~\ref{fig:reprojection_error_comparison} shows reprojection error vectors for the proposed line intersection approach and \emph{libcbdetect}.
Vector heads are connected into a grid representing the boards and revealing the nature of the error.
As we can see the default \emph{libcbdetect} is more noisy compared to the proposed line intersection which has very uniform grids with minimal noise.
From Table  \ref{tbl:rep_err_compare}, we can see that the proposed method has smaller reprojection errors than \emph{libcbdetect} and OpenCV's method.
The remaining reprojection error of approximately 0.1 pixel is systematic and cannot be solved with more precise corner detection.
For example, board 11 (second to last on the right) is twisted since it is quite long, flexible, and leaned against the wall.
Thus, in reality it is not a plane, and fitting a plane model to this subset of points introduces an error much larger than the other boards.
For reference, we also provide reprojection error of our method without this board in the last column of Table \ref{tbl:rep_err_compare}.
We believe that the remaining reprojection error comes from the discrepancy between radial-tangential model and the distortion function of the real lens.

\subsection{Board matching and optimization}

The primary goal of this paper is to investigate if another set of calibration parameters exists for the KITTI dataset  that can produce more accurate trajectory for a general visual odometry algorithm.
Therefore, we did not aim to design a general robust calibration method as long as simple concepts were successful on the KITTI images.
Given that, we assume that the number of boards in each image is equal.
Then, in each image, the detected boards are sorted from left to right.
Following this order, each board in the left image gets associated with the first free board from the right image with an equal number of rows and columns.
Since the corners inside one board are always ordered in the same snake-like pattern starting from the top-left corner, correct board-to-board match also implies correct corner-to-corner match.
We assume a pinhole camera model with radial-tangential distortion yielding a total of $9$ intrinsic parameters per camera $(f_x,f_y,c_u,c_v,k_1,k_2,k_3,k_4,k_5)$, i.e., focal lengths, principal point, and lens distortion parameters, and $6$ extrinsic parameters for each camera except the referent one.
We initialize the focal length with some value close to the default KITTI solution, while we set the principal point with the image center and distortion parameters to zero.
The extrinsic parameters were initialized by averaging the checkerboard-to-image plane homographies \cite{Zhang1999}. Finally, the sum of squared reprojection errors is minimized with the Levenberg-Marquard algorithm.
We did not notice any instability of the $k_5$ parameter as reported in \cite{Geiger2012}.

\subsection{Calibration parameters refinement}

Following steps described in previous sections, we obtained calibration parameters with slightly more accurate odometry compared to \emph{libcbdetect}, but the result was still below our expectations.
Given that, we decided to refine a subset of calibration parameters, i.e., we focused on 4 parameters for which board configurations did not strongly constrain the problem.

\begin{figure}[!t]
	\begin{center}
		\begin{subfigure}[b]{0.93\hsize}
			\centering
			\includegraphics[width=\hsize]{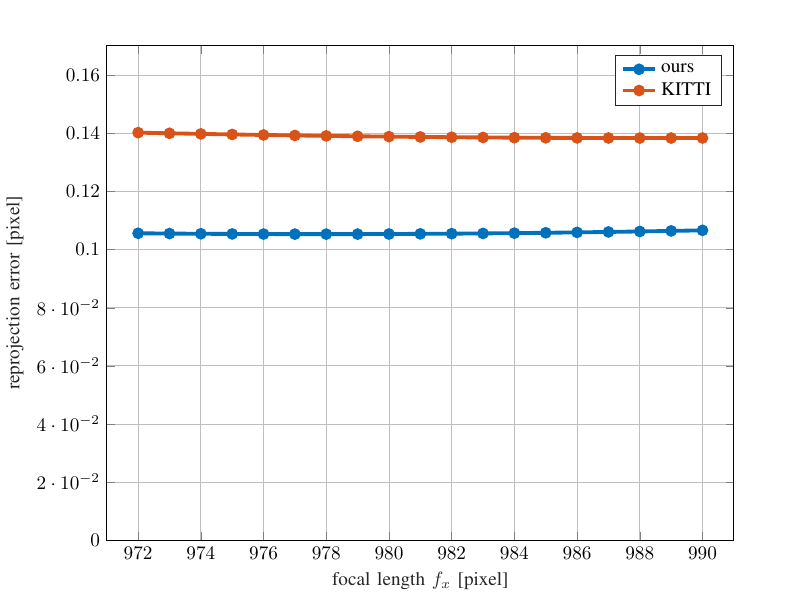}
			\vspace{-0.7cm}
			\caption{}
			\label{fig:error_fx}
			\vspace{-0.3cm}
		\end{subfigure}\\
		\begin{subfigure}[b]{0.93\hsize}
			\centering
			\includegraphics[width=\hsize]{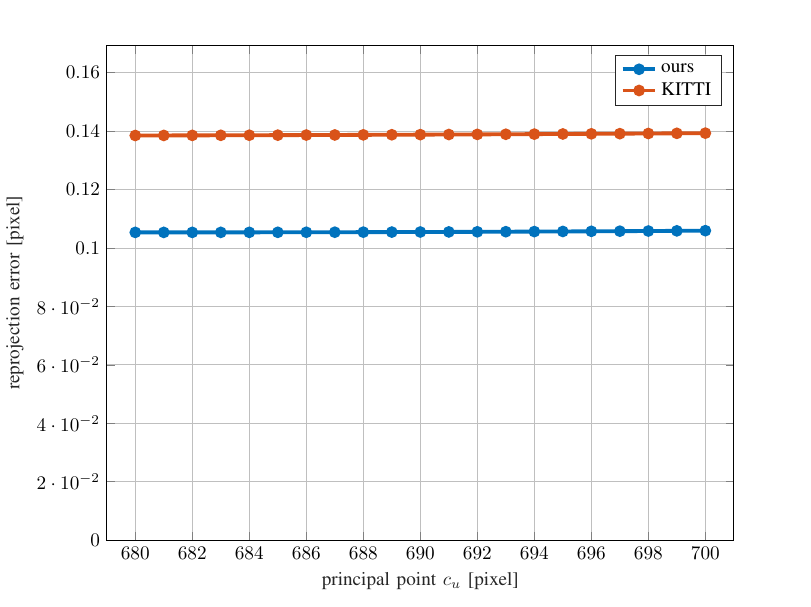}
						\vspace{-0.7cm}
			\caption{}
			\label{fig:error_cu}
		\end{subfigure}
		\caption{Reprojection error with respect to left camera (a) focal length $f_x$ and (b) principal point $c_u$. Both have flat error curves indicating low sensitivity of the optimization criterion to changes in these parameters. The same effect applies to $c_v$.}
								\vspace{-0.7cm}
		\label{fig:error_rot}
	\end{center}
\end{figure}

In Fig.~\ref{fig:error_rot} we show the reprojection error  with respect to focal length $f_x$ and the principal point coordinate $c_u$ of the left camera.
We conducted sensitivity analysis by fixing the parameter to a specific value and then conducted the calibration optimization.
The parameters were sampled around the initial minimum value with one pixel steps to the left and to the right.
While the change in the focal length and principal point significantly influence the odometry accuracy, this is not the case with the calibration reprojection error, as we can see in Fig.~\ref{fig:error_rot}.
The reprojection error curves are nearly flat relative to the focal length and principal point.
Values 10 pixels away from the minimum point have an error slightly smaller than 0.001 pixels which is below our estimated system noise of approximately  0.03 pixels that is due to the influence of the bent board or the difference between two corner refinement methods (see Table~\ref{tbl:rep_err_compare}).
We concluded that the provided board configurations do not constrain enough the calibration for accurate estimation of all the parameters.
Therefore, we decided to change the objective function and search for the focal length and principal point that minimize the rotational part of the error in the KITTI evaluation metric.

We employed grid search around the initial minimum such that the probed values were fixed during the calibration process.
Then, the odometry with obtained calibration parameters was evaluated for the rotational error on the training sequences and best parameters were found within the neighborhood of 10 pixels from the initial value.
While the calibration reprojection error increased for approximately $1\%$, odometry error was reduced up to $50\%$.
Note that in the calibration process, the focal length and principal point were fixed for the left camera only, while the corresponding parameters in the right camera are highly correlated via boards and they automatically change with the probed left camera values.
Also, there is a strong correlation between the focal length in the $x$ and $y$ direction and refinement in one of them is instantly reflected in the other one.
Therefore, while minimizing the rotation error, only three parameters are obtained using grid search, namely $f_x$ and $(c_u, c_v)$ of the left camera.

\begin{figure}[!t]
	\centering
	\includegraphics[width=0.93\columnwidth]{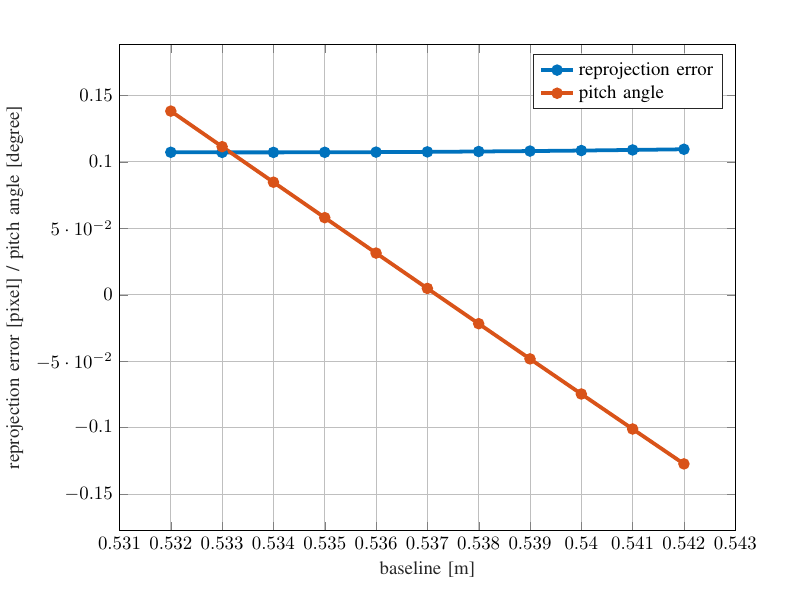}
	\caption{Reprojection error (blue) and the relative angle between the cameras with respect to the baseline (red).}
	\vspace{-0.7cm}
	\label{fig:baseline}
\end{figure}

In the next step we focused on the scale which mostly depends on the estimated stereo baseline.
Note that if the odometry algorithm estimates rotation and translation jointly, then baseline should be probed along with the first three parameters.
Fig.~\ref{fig:baseline} shows the reprojection error relative to the baseline around the initial minimum value.
Again, we can notice that in this case the reprojection error curve is flat.
However, there is a strong correlation between the baseline and relative angle between the cameras.
Forcing the baseline to a different value results in the change of the camera pitch angle around $y$ axis, while retaining almost the same reprojection error (note that local camera coordinate system is right-handed with the $z$ axis pointing in the direction of the optical axis).
In a way, the system cannot disambiguate between the baseline and the pitch angle near the exact solution, since different triangulation setups lead to similar reprojection error.
Consequently, exact baseline is hard to assess with odometry, since the trajectory scale may also depend on the pitch angle.
In the end, a total number of 4 parameters is refined via grid search: $f_x$ and $(c_u, c_v)$ of the left camera, and the baseline.
All other parameters change accordingly during the calibration optimization process since they are all closely correlated.
The KITTI calibration file containing all the calibration parameters that were used in this paper is publicly available\footnote{https://bitbucket.org/unizg-fer-lamor/kittical}.

\section{Experimental results}

We demonstrate the effectiveness of the obtained calibration parameters on three different visual odometries: SOFT2 \cite{Cvisic2021a}, ORB-SLAM2 \cite{Mur-Artal2016} and VISO2 \cite{Geiger2011}.
These algorithms are also different in nature: SOFT2 minimizes point-to-line (2D-2D) distances and uses left camera only for all motion parameters except for the scale, ORB-SLAM2 maps 3D points and refines their positions via local bundle adjustment, while VISO2 is pure frame-to-frame odometry  minimizing reprojection errors of 3D points projected from a single disparity frame.
Since our method requires odometry to refine the subset of calibration parameters, we note that this can introduce a bias towards that particular method.
Nevertheless, our new calibration parameters improved each of the three algorithms.
In the presented experiment we used ORB-SLAM2 for parameter refinement.
Table \ref{tbl:kitti_compare} shows the results with parameters seeded from $f_x=975$ pixels, $(c_u, c_v) = (700, 247)$ pixels, and $\mathrm{baseline} = 0.539$\,m for SOFT2, ORB-SLAM2, and VISO2, for each training sequence along with the overall average result.
To put emphasis on pure odometry drift, loop closing feature of ORB-SLAM2 was turned off.
The experiment was conducted on raw KITTI sequences, where images were prerectified with new parameters before they were used in the odometry.
Each odometry showed notable improvement on majority of the tracks, and on average  $30\%$ improvement in translational error and $50\%$ improvement in rotational error.

\begin{table*}[!t]
	\caption{SOFT2, ORB-SLAM2, and VISO2 results with default and custom calibration for $10$ KITTI dataset train sequences (d/hm stands for degrees/100 meters)}
	\label{tbl:kitti_compare}
	\begin{center}
		\begin{tabular}{ |c|c|c|c|c|c|c|c|c|c|c|c|c|  }
			\hline
			& \multicolumn{2}{|c|}{SOFT2 default} & \multicolumn{2}{|c|}{SOFT2 recal.} & \multicolumn{2}{|c|}{ORB-SLAM2 default} & \multicolumn{2}{|c|}{ORB-SLAM2 recal.} & \multicolumn{2}{|c|}{VISO2 default} & \multicolumn{2}{|c|}{VISO2 recal.} \\
			\hline
			Seq. & $t_{\text{rel}} [\%]$ & $r_{\text{rel}}$ [d/hm] & $t_{\text{rel}} [\%]$ & $r_{\text{rel}}$ [d/hm] & $t_{\text{rel}} [\%]$ & $r_{\text{rel}}$ [d/hm] & $t_{\text{rel}} [\%]$ & $r_{\text{rel}}$ [d/hm] & $t_{\text{rel}} [\%]$ & $r_{\text{rel}}$ [d/hm] & $t_{\text{rel}} [\%]$ & $r_{\text{rel}}$ [d/hm]\\
			\hline
			00 & 0.68 & 0.28 & \textbf{0.47} & \textbf{0.16} & 0.88 & 0.31 & \textbf{0.57} & \textbf{0.17} & 1.53 & 0.64 & \textbf{0.89} & \textbf{0.33} \\
			01 & 1.07 & 0.16 & \textbf{0.73} & \textbf{0.08} & 1.44 & \textbf{0.19} & \textbf{1.23} & 0.21 & \textbf{3.80} & 0.70 & 3.89 & \textbf{0.30} \\
			02 & 0.73 & 0.21 & \textbf{0.50} & \textbf{0.11} & 0.77 & 0.28 & \textbf{0.54} & \textbf{0.16} & 1.61 & 0.52 & \textbf{0.79} & \textbf{0.24} \\
			04 & 0.55 & 0.07 & \textbf{0.30} & \textbf{0.05} & 0.46 & \textbf{0.19} & \textbf{0.42} & 0.22 & 1.29 & 0.51 & \textbf{0.93} & \textbf{0.45} \\
			05 & 0.61 & 0.24 & \textbf{0.38} & \textbf{0.11} & 0.62 & 0.26 & \textbf{0.56} & \textbf{0.13} & 1.25 & 0.66 & \textbf{0.80} & \textbf{0.26} \\
			06 & 0.60 & 0.23 & \textbf{0.38} & \textbf{0.12} & 0.80 & 0.25 & \textbf{0.39} & \textbf{0.11} & \textbf{0.79} & 0.51 & 0.86 & \textbf{0.30} \\
			07 & 0.50 & 0.33 & \textbf{0.30} & \textbf{0.16} & 0.89 & 0.50 & \textbf{0.43} & \textbf{0.20} & 1.46 & 1.13 & \textbf{0.73} & \textbf{0.51} \\
			08 & 1.00 & 0.28 & \textbf{0.80} & \textbf{0.16} & 1.03 & 0.31 & \textbf{0.86} & \textbf{0.17} & 1.62 & 0.66 & \textbf{1.30} & \textbf{0.41} \\
			09 & 0.75 & 0.18 & \textbf{0.59} & \textbf{0.08} & 0.86 & 0.25 & \textbf{0.75} & \textbf{0.15} & \textbf{0.84} & 0.64 & 1.46 & \textbf{0.42} \\
			10 & 0.65 & 0.24 & \textbf{0.58} & \textbf{0.14} & 0.62 & 0.29 & \textbf{0.58} & \textbf{0.23} & 1.29 & 0.64 & \textbf{1.01} & \textbf{0.43} \\
			\hline
			avg& 0.76 & 0.24 & \textbf{0.55} & \textbf{0.13} & 0.87 & 0.29 & \textbf{0.65} & \textbf{0.16} & 1.62 & 0.63 & \textbf{1.11} & \textbf{0.33} \\
			\hline
		\end{tabular}
	\end{center}
\end{table*}

However, results for SOFT2 on the KITTI scoreboard could not be obtained with raw images, since they are not provided for the testing subset.
Nevertheless, since KITTI default intrinsic and extrinsic parameters are given, we have implemented a coordinate converter after the feature detection and matching block.
Each feature 2D coordinate is distorted to the original image via default KITTI parameters, and then undistorted with custom parameters before the optimization.

\section{Conclusion}
In this paper we have presented a novel approach to one shot calibration of the KITTI dataset camera setup.
The approach is based on detecting line segments for sub-pixel corner accuracy and board matching for final parameter estimation, finally outperforming the default \emph{libcbdetect} and OpenCV's calibration.
We tested new calibration parameters on three methodologically different visual odometry algorithms, namely SOFT2, ORB-SLAM2, and VISO2, and all the three showed improved accuracy.
Finally, with the proposed method our SOFT2 achieved the highest accuracy on the official KITTI scoreboard with 0.53\% translational and 0.0009 deg/m rotational error, outperforming even 3D laser-based methods.
Although we are targeting a specific dataset, we believe that the results are relevant to the community since the paper offers insights of general interest and KITTI will probably remain being a reference dataset in the forthcoming years.

%

%

%
%

\ifCLASSOPTIONcaptionsoff
  \newpage
\fi




\balance
\bibliographystyle{IEEEtran}
\bibliography{library,igor_bib}

%

%






\end{document}